\documentclass[journal,twoside,web]{IEEEtran}

\usepackage{hyperref}
\usepackage{graphicx}
\usepackage{multirow}
\usepackage{amssymb}
\usepackage{amsmath}
\usepackage{booktabs}
\usepackage[font=small, labelfont=bf]{subcaption}
\usepackage{algpseudocode}
\usepackage[ruled,vlined]{algorithm2e}
\usepackage{colortbl}
\usepackage{xcolor}
\usepackage{soul}

\newcommand{\beq}{\begin{equation}}
\newcommand{\eeq}{\end{equation}}

\newcommand{\Real}{\mathbb{R}}

\newcommand{\omitme}[1]{}

\newcommand{\figref}[1]{Fig.~\ref{#1}}
\newcommand{\confusedHL}[1]{}
\newcommand{\confusedJM}[1]{}

\begin{document}
\title{Attention-based Dynamic Subspace Learners for~Medical~Image~Analysis}

%


\author{Sukesh Adiga V, \and 
        Jose Dolz, \and
        and Herve Lombaert  \\ 
\thanks{All authors are with École de technologie supérieure (ETS) Montreal, Canada. (\textit{Corresponding author: Sukesh Adiga V,  \textbf{Email:}~{sukesh.adiga-vasudeva.1@ens.etsmtl.ca}.)}}
}


%

\maketitle              


\begin{abstract}
Learning similarity is a key aspect in medical image analysis, particularly in recommendation systems or in uncovering the interpretation of anatomical data in images. Most existing methods learn such similarities in the embedding space over image sets using a single metric learner. Images, however, have a variety of object attributes such as color, shape, or artifacts. Encoding such attributes using a single metric learner is inadequate and may fail to generalize. Instead, multiple learners could focus on separate aspects of these attributes in subspaces of an overarching embedding. This, however, implies the number of learners to be found empirically for each new dataset. This work, Dynamic Subspace Learners, proposes to dynamically exploit multiple learners by removing the need of knowing \textit{apriori} the number of learners and aggregating new subspace learners during training. Furthermore, the visual interpretability of such subspace learning is enforced by integrating an attention module into our method. This integrated attention mechanism provides a visual insight of discriminative image features that contribute to the clustering of image sets and a visual explanation of the embedding features. The benefits of our attention-based dynamic subspace learners are evaluated in the application of image clustering, image retrieval, and weakly supervised segmentation. Our method achieves competitive results with the performances of multiple learners baselines and significantly outperforms the classification network in terms of clustering and retrieval scores on three different public benchmark datasets. Moreover, our method also provides an attention map generated directly during inference to illustrate the visual interpretability of the embedding features. These attention maps offer a proxy-labels, which improves the segmentation accuracy up to 15\% in Dice scores when compared to state-of-the-art interpretation techniques.
\end{abstract}


\section{Introduction}
Learning the similarity between arbitrary images is a fundamental problem in many key areas of computer vision such as image retrieval \cite{sohn2016improved,movshovitz2017no,he2018triplet}, recommender system \cite{ma2019learning,liu2019single}, duplicate detection \cite{zheng2016improving}, clustering \cite{ziko2018scalable}, or zero-shot learning \cite{zhang2016zero}. In this context, metric learning is commonly used for measuring similarities by learning a distance function over objects \cite{weinberger2006distance,kulis2012metric}. Recently, deep metric learning (DML) has been raised as a powerful approach to learn these similarities \cite{kaya2019deep}. More specifically, the goal of DML is to learn an embedding space where images from the same classes are encouraged to be close to one another. In contrast, images belonging to different classes are pushed away in the embedding space. In recent DML approaches, the loss function can be typically expressed in Euclidean distances or cosine similarities between pairs or tuples of images in the embedding space. Well-known losses employed in DML include: contrastive loss \cite{hadsell2006dimensionality}, triplet loss \cite{wang2014learning}, lifted structure loss \cite{oh2016deep}, N-pairs loss \cite{sohn2016improved}, margin loss \cite{wu2017sampling}, angular loss \cite{wang2017deep}, or ProxyNCA loss \cite{movshovitz2017no}. In addition to novel learning objectives, recent efforts are also devoted to designing efficient sample-mining \cite{wu2017sampling}, or sample weighting \cite{wang2019multi} strategies.

Most of these methods learn the embedding mapping function with a single metric learner. However, medical images have complex distributions consisting of different object attributes such as color, shape, size, or artifacts. Thus, learning the complex similarity associated with these different object attributes may be inadequate with only one single-learner. A few attempts have been made towards leveraging multiple metric learners to address this complexity. For example, Kim \textit{et al.} \cite{kim2018attention} ensemble multiple learners, whereas a \textit{divide-and-conquer} strategy is used in \cite{sanakoyeu2019divide} by splitting the manifold into several embedding subspaces. One main limitation of these approaches is a need to empirically find the optimal number of learners, which requires a new validation for every new setting, including every use of a new dataset. Furthermore, the sizes of the embedding subspaces associated with each learner might differ since learning the various sets of object attributes requires varying degrees of modeling complexity.


Despite the popularity of DML, surprisingly few works attempt to visually explain which regions contribute to the similarity between images in embedding networks \cite{hu2022x}. These visualizations are of pivotal importance since they provide an efficient mechanism to understand the predictions of the model. Recent efforts have been devoted to the interpretability of deep neural networks, resulting in a variety of different approaches \cite{zeiler2014visualizing,koh2017understanding,selvaraju2017grad,chen2019explaining,belharbi2020deep}. Among these methods, GradCAM \cite{selvaraju2017grad} has been widely employed to explain deep classification models. This method uses gradients to highlight the discriminative regions of an image. Nevertheless, since the gradients are not available during testing, directly applying this strategy in embedding networks is not feasible \cite{chen2020adapting}. Integrating interpretability in embedding networks requires either attaching an additional classification branch \cite{zheng2019re} or employing multiple images simultaneously \cite{stylianou2019visualizing,zhu2021visual}. Needless to say, interpretability is of particular interest in medical imaging, as visual explanations of predictions directly impact the diagnosis, therapy planning, and follow-up of many diseases. Thus, existing DML approaches may be inadequate to visually uncover what constitutes similarities among a complex set of medical images.

Motivated by these gaps and the scarcity of the DML literature in medical imaging, we propose a novel attention-based dynamic subspace learners approach. The underlying metric learning method is inspired by the idea of a \textit{divide-and-conquer} strategy. More specifically, we propose to follow the approach of \cite{sanakoyeu2019divide} in order to capture different object attributes, each of them processed with an independent subspace learner. These subspace learners having variable sizes are learned dynamically as and when the network accuracy is plateauing during training. Thereby avoids the need to find \textit{apriori} the number of subspace learners while retaining the state-of-the-art performance. Furthermore, the visual interpretation of the embedding is addressed by integrating an attention module after feature extraction layers, encouraging the learners to focus on the discriminative areas of target objects. Consequently, the learning process provides a visual insight of which image region considerably contributes to the clustering of image sets in the form of pixel-wise interpretable predictions.




\subsubsection*{Our Contribution}
We contribute a novel approach to the state-of-the-art method in deep metric learning and illustrate its application in medical image analysis. More precisely, we propose a training strategy that (i) explores the dynamic learning of an embedding, (ii) overcomes the empirical search of an optimal number of subspaces in approaches based on multiple metric learners, and (iii) produces compact subspaces of variable size to attend different object attributes. Furthermore, the integration of an attention module in our dynamic learner approach focuses the attention of each independent learner on the discriminative regions of an object of interest. This attention mechanism provides the added benefit of visually interpreting relevant embedded features. The evaluation of our proposed method is conducted by extensive experiments on three publicly available benchmarks: ISIC19 \cite{codella2019skin,combalia2019bcn20000}, MURA \cite{rajpurkar2017mura}, and HyperKvasir \cite{borgli2020hyperkvasir}. The performance is evaluated on clustering and image retrieval tasks, showing that the proposed method achieves competitive results with the state-of-the-art without requiring the grid searches over optimal numbers of learners. We also demonstrate that the attention maps produced by our method can be used as proxy-labels to train deep segmentation models. In particular, we evaluate our approach on ISIC18 \cite{tschandl2018ham10000,codella2019skin} in a weakly supervised segmentation task and show improvements to the visual attention and class activation maps obtained from recent state-of-the-art methods, including the method specifically designed for skin lesion detection \cite{zhang2019attention}. 


\section{Related Work}

\subsection{Deep Metric Learning}
Metric learning is a widely explored research field in the learning community \cite{bromley1994signature,weinberger2006distance}. The seminal work of Siamese Networks \cite{bromley1994signature} represents the first attempt to use neural networks for feature embedding. Its concept is to employ two identical neural networks that learn a contrastive embedding from a pair of images. With the advent of deep learning, deep metric learning (DML) has gained popularity, becoming a mainstay in many modern computer vision problems, such as image retrieval \cite{opitz2017bier}, person re-identification \cite{liao2015person}, or few-shot learning \cite{snell2017prototypical}. In DML, the images are mapped into a manifold space via deep neural networks. Euclidean or cosine distances can then be directly used as a metric distance between two images in this mapped space. Typical losses employed in DML include contrastive \cite{hadsell2006dimensionality} or triplet loss \cite{schroff2015facenet}. The contrastive loss \cite{hadsell2006dimensionality} encourages images from the same class to stay closer --in the learned manifold-- while pushing away samples from different classes, which should be separated by a given fixed distance. Nevertheless, forcing the same distance for all pairs of images can discourage any potential distortion in the embedded space. In contrast, this assumption is relaxed in triplet loss \cite{hadsell2006dimensionality}, which only imposes that negative pairs of images should be further away than positive pairs.

In the same direction as our work, \cite{kim2018attention} and \cite{sanakoyeu2019divide} have leveraged the use of multiple learners to diversify the learning space towards different object attributes. While \cite{kim2018attention} propose an ensemble of multiple learners driven by attention, a \textit{divide and conquer} strategy is employed in \cite{sanakoyeu2019divide}, which promotes the discovery of multiple subspaces. For example, Sanakoyeu \textit{et al.} \cite{sanakoyeu2019divide} explicitly splits the embedded space into a predefined number of learners with fixed size subspaces. Then, each learner independently learns a part of an embedding space, i.e., a subspace, from a portion of clustered data, and the final embedding is later refined from multiple learners. Even though this strategy leads to improvements over its single-learner counterpart, a grid search is needed to find an optimal number of learners with each new dataset. Furthermore, the size of the embedding space is uniform across the learners, whereas some attributes, such as color, might require smaller embeddings to encode the information than other attributes, such as shape.



\subsection{Metric Learning in Medical Image Analysis}
Despite the interest in other domains, metric learning, and more particularly DML, remains almost unexplored in medical imaging. In the pre-deep learning era, related work includes \cite{yang2008boosting}, which employed a distance metric learning in a traditional boosting framework in a medical image retrieval scenario. More recently, \cite{yan2018deep} investigates the use of DML to model the similarity relationship between lesions in the context of radiology images, where a triplet loss is employed to learn the lesion embeddings. Gupta \textit{et al.} \cite{gupta2019deep} also resorts to the triplet loss to learn the underlying manifold space for the task of Mitotic classification, whose embedded features are subsequently used as input for a Support Vector Machine classifier. Recently, a combination of cross-entropy loss with a contrastive loss or triplet loss is used to classify whole slide images in digital pathology \cite{teh2019metric,pati2020reducing}. In \cite{sikaroudi2020supervision}, a triplet loss is used to learn a representation of source domain images, which is later used for target domain classification under the few-shot learning paradigm. In \cite{teh2020learning}, DML is used to pre-train a model in the application of digital pathology classification, where authors use a ProxyNCA loss for learning transferable features. To enhance the embedding, \cite{zhong2021deep,yang2019liver} has integrated a multi-similarity loss to DML in the context of chest radiography and liver histopathology image, respectively.
Nevertheless, most of these methods are developed with the goal of classification tasks and do not effectively leverage the geometrical information of the underlying embedding space.

\subsection{Weakly Supervised Segmentation}
Weakly supervised segmentation (WSS) has emerged as an alternative to alleviate the need for large amounts of \textit{pixel-level} labelled data. These labels can come in the form of \textit{image-level} labels \cite{papandreou2015weakly}, scribbles \cite{lin2016scribblesup}, points \cite{bearman2016s}, bounding boxes \cite{rajchl2016deepcut} or direct losses \cite{kervadec2019constrained}. Among them, image-level labels are easier and inexpensive to obtain \cite{bearman2016s}. 
Particularly, class activation maps (CAM) \cite{zhou2016learning} have gained popularity in identifying saliency regions based on image labels. It is achieved by associating feature maps of the last layers and weighting their activation using a global average pooling (GAP) layer. However, generated saliency maps are typically spread around the target object, only focusing on the most discriminant areas. This limits its usability as pixel-level supervision for semantic segmentation. To enhance the generated saliency regions, some alternatives based on back-propagation (GradCAM \cite{selvaraju2017grad}) or super-pixels (SP-CAM \cite{kwak2017weakly}) have been proposed. Nevertheless, these methods demand additional gradients computations \cite{selvaraju2017grad} or supervisions \cite{kwak2017weakly}.

The literature on WSS in medical imaging remains scarce.
While few methods resort to direct losses, hence requiring additional priors, such as the target size \cite{jia2017constrained,kervadec2019constrained}, other approaches rely on stronger forms of supervision, for instance, using bounding boxes \cite{rajchl2016deepcut} or scribbles \cite{can2018learning}. Tackling WSS from a perspective of \textit{image-level} labels typically involves visual features, which has not been thoroughly investigated \cite{feng2017discriminative,meng2019weakly,nguyen2019novel,dubost2020weakly}. For example, Nguyen \textit{et al.} \cite{nguyen2019novel} has proposed a CAM-based approach for the segmentation of uveal melanoma. In their method, the CAMs generated by the classification network are further refined by an active shape model and conditional random fields \cite{krahenbuhl2011efficient}. More recently, CAMs derived from image-level labels have been combined with attention scores to refine lesion segmentation in brain images \cite{wu2019weakly}. By doing so, they have demonstrated a performance improvement compared to the vanilla version of CAMs. Nevertheless, these methods typically integrate CAM/GradCAM with complex models to enhance the performance of a final segmentation.

\section{Methodology}
\label{sec:methods}

\begin{figure*}[th!]
\centering
\includegraphics[width=0.985\linewidth]{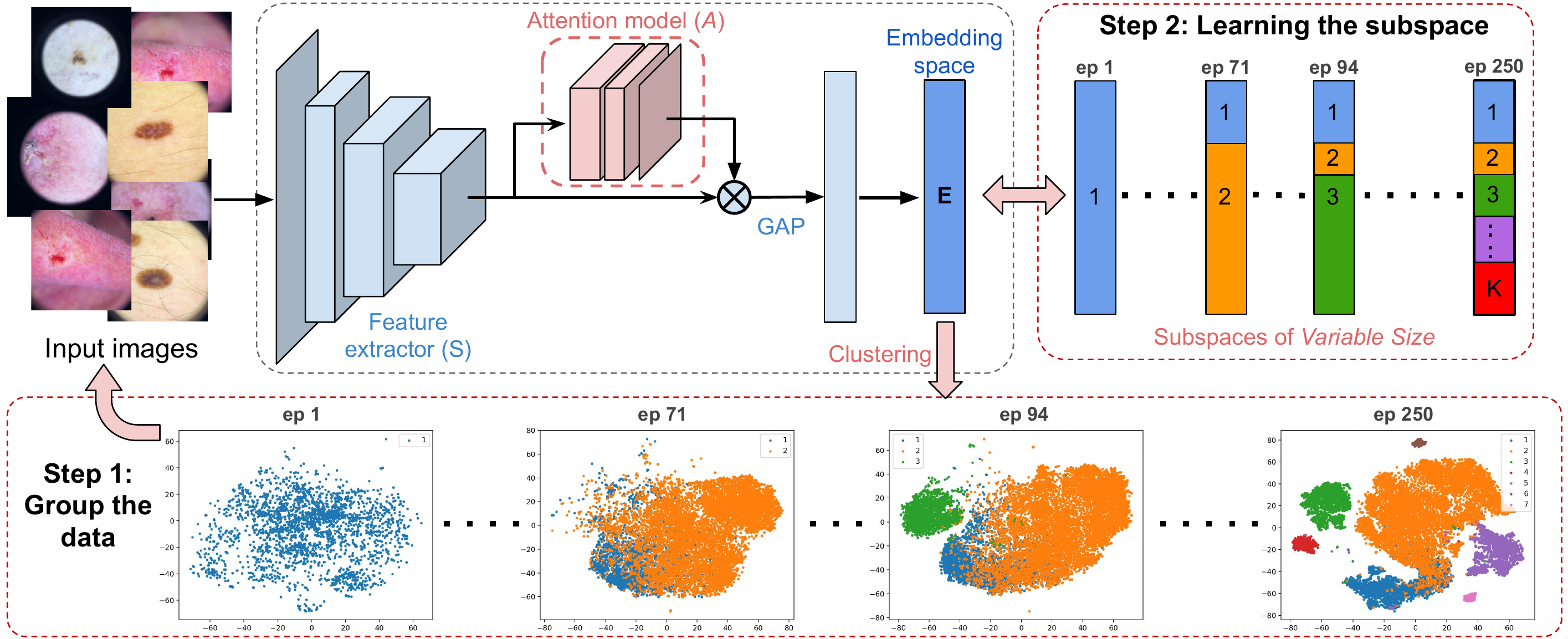}
\caption{\textbf{Overview of our proposed attention-based dynamic subspace learners} - The embedding space is dynamically divided into the subspaces of varying sizes during training. Suppose there are $K$ subspaces at a particular training time; the data are first grouped into $K$ groups in the full embedding space (step 1, from epoch 1 to 250) and assign each subgroup of data to an individual subspace learner. Each learner then only attends the data from its subgroup in the learning stage (step 2). In inference time, our method uses the entire embedding space $E$ to map an image. Best viewed in color.}
\label{fig:arch}
\end{figure*}

\subsection{Overview}
An overview of the proposed approach is depicted in \figref{fig:arch}. The main idea is to split the embedding space into multiple subspaces ($K$) such that the original embedding space can be learned by refining its subspaces. Contrary to \cite{sanakoyeu2019divide}, the embedding space is split dynamically, which removes the need to search for the optimal number of learners $K$ in each scenario. The whole process is divided into two iterative steps. First, input images are mapped into the lower dimension embedding space using the entire embedding layer $E$ ($d$-dimension), where they are clustered into different groups. Second, the clustered data is consequently assigned to an individual subspace learner, where their corresponding images are used to train each subspace. These two steps are repeated at regular intervals, as well as each time a new learner is added. The key idea is that each subspace learner learns a part of the embedding space from a subgroup of images instead of learning a whole embedded representation vector. Finally, all subspaces are combined to generate a full embedding space. Furthermore, an attention module is integrated within the learning process to guide the learning of distance metrics. The following sections describe the deep metric learning formulation, present the proposed dynamic subspace metric learning and attention module.



\subsection{Deep Metric learning Formulation}

Let the training dataset be defined as $\textbf{X} = \{(x_i,y_i)\}^N_{i=1}$, where the \textit{i}-th image is denoted as $x_i \in \Real^m$, and $y_i \in \{1,2,...,C\}$ is its corresponding class label. $C$ defines the total number of classes. The goal of deep metric learning is to learn an embedding function $f_\theta(\cdot): \Real^m \rightarrow \Real^d$, which discriminatively maps semantically similar images (same class) in the input space $\Real^m$ onto metrically close points in the learned manifold $\Real^d$. Similarly, semantically dissimilar images (different class) in $\Real^m$ should be mapped metrically far in $\Real^d$. The parameters $\theta$ of the mapping function are typically learned by a convolutional neural network. Formally, the distance metric $d(x_i, x_j): \Real^d \times \Real^d \rightarrow \Real$ between two images in the embedding space $\Real^d$ can be defined as:

\beq
\label{eq:dist}
d(x_i, x_j) = || f_\theta(x_i) - f_\theta(x_j) ||,
\eeq
where $||\cdot||$ denotes the Euclidean norm. This distance can be minimized in different ways, depending on the loss function employed. In this work, we resort to the Margin loss \cite{wu2017sampling}:

\beq
\label{eq:margin}
l_{margin}(x_i, x_j) = [\alpha + \mu_{ij} (d(x_i, x_j) - \beta)]_+,
\eeq
where $\beta$ is the boundary between the similar and dissimilar pairs, $\alpha$ is a separation margin, and $\mu_{ij} \in \{-1, 1\}$ indicates whether the images in the pair are similar ($\mu_{ij}=1$) or different ($\mu_{ij}=-1$). Note that any other metric learning loss function can be employed with our approach.



\subsection{Dynamic Subspace Learners}
The complexity of the original problem can be solved by dividing the problem into smaller sub-problems, which are easier to solve. We follow the approach in \cite{sanakoyeu2019divide}, where the embedding space $\Real^d$ and the data is split into multiple groups. Specifically, splitting of the embedding space is conducted by slicing the $\Real^d$ space, i.e., the last dense layer of the network, into $K$ sub-vectors of the same size, $d/K$. Furthermore, data is clustered into $K$ groups based on their pairwise distance in the embedding space $\Real^d$, for instance, using K-means. Then, a set of $K$ independent learners is used to learn over each subspace by using a fraction of the input data, thereby reducing the complexity of the original problem. Nevertheless, a major bottleneck is finding an optimal number of subspaces $K$ to learn an effective embedding, which must be found empirically for every new dataset. Moreover, the subspace is divided equally, which is ineffective as not all the object attributes require the same size to encode the information.

Contrary to \cite{sanakoyeu2019divide}, our proposed learning strategy finds an optimal embedding by dynamically splitting the embedding space and associating with a metric learner during training.
To construct each subspace, we group highly contributing neurons of the embedding layer $E$, which is repeated until network convergence.
Initially, the entire embedding space is learned with all the data, with an initial single learner $K = 1$. As the learning progresses, the accuracy of the model starts to reach an initial plateau. At this stage, we compute the score of each neuron ($e_i$) in the embedding layer, similarly to the pruning strategy as in \cite{molchanov2016pruning}. In particular, the low-scoring neurons are pruned such that the performance drop of the model is minimal, i.e, $ \left| \Delta f_{\theta}(e_{i}) \right| =  \left| f_{\theta}(X, e_i = 0) - f_{\theta}(X, e_i) \right| $. By using Taylor expansion, as in \cite{molchanov2016pruning}, the scoring of each neuron $e_i$ can be reduced to:

\beq
s(e_i) = \left| \Delta f_{\theta}(e_{i}) \right| = \left| f_{\theta}(X, e_i) - \frac{\partial f_{\theta}}{\partial e_i} e_i  - f_{\theta}(X, e_i) \right| = \left| \frac{\partial f_{\theta}}{\partial e_i} e_i \right|
\label{eq:score}
\eeq


Thus, the scoring of neurons is simplified to multiplying the activation and the gradient output in the embedding layer. This score $s(e_i)$ is computed for each training example separately, and is consequently averaged across all training data and normalized to $[ 0, 1 ]$. The neurons having high normalized scores are subsequently grouped to form a new subspace. Particularly, the neurons having more than 50\% of the confidence score, i.e., $s(e_i) > 0.5$, are grouped as a new subspace. The current metric learner ($\mathcal{L}_{k}$) is later assigned to this group of neurons. The remaining neurons of the embedding layer, $e_{r}$, are eventually reset, similar to the pruning technique \cite{molchanov2016pruning} and assign a new metric learner as in Eq.~\ref{eq:learner}. After adding this new learner, the training data is clustered by mapping into the entire embedding space using K-means with the updated $K$ ($K=2$ for the second iteration). Note that the entire embedding space here is a combination of all the subspaces. Each learner is eventually assigned a subgroup of data from the clustering, resulting in each learner being trained with a fraction of the input data. The addition of a new learner is repeated with the remaining neurons $e_{r}$ when the network performance reaches a new plateau, until convergence. In the end, it results in $K$ mapping functions, $\textbf{f}=[f^1,f^2,...,f^{K}]$, where each mapping function $f^k$ will project the images $\Real^m$ into the corresponding subspace of $\Real^{d_k}$, each with a variable size.



All learners are trained jointly by resorting to the margin loss \cite{wu2017sampling}, which for each learner can be defined as: 

\beq
\label{eq:learner}
\mathcal{L}_{k}^{f^k_{\theta_k}}(x_i, x_j) = \sum_{(x_i, x_j)\sim B} [\alpha + \mu_{ij} (d_{f^k_{\theta_k}}(x_i, x_j) - \beta)]_+,
\eeq

where $(x_i, x_j)\sim B$ is the current mini-batch (uniformly sampled from each data group) having both positive and negative classes, and $d_{f^k_{\theta_k}}$ is the distance metric (similar to Eq.\ref{eq:dist}) for the $k$-th learner. Once individual learners are trained, these are merged to compose the entire embedding space, which is refined with the entire training set. Furthermore, assuming that the learned embedding space is improving over time, we re-cluster the images at every $T_c$ epochs by mapping all the images using the entire embedding space $E$. An outline of the proposed method is presented in Algorithm~\ref{algo:DynK}.


\begin{algorithm}[t]
\small
\DontPrintSemicolon
\vspace{1mm}
\SetKwInOut{Parameter}{Inputs}
\Parameter{$X$, $X_{test}$ : Training and test data \\
    $\theta$ : backbone network parameters \\
    E : Embedding space \\
    $T_c$, $T_p$ : clustering and network plateau threshold}
\vspace{1mm}
\SetKwInOut{Parameter}{Initialize}
\Parameter{K $\leftarrow$ 1, number of learner \\
$B \leftarrow$ 0, Best epoch \\
ep $\leftarrow$ 1, current epoch \\
$e_r \leftarrow E$, remaining embedding space \\
RC $\leftarrow$ True, re-clustering flag}

\vspace{1mm}
\While{Not converged}{
    \If(\Comment{Re-cluster the data}){RC}{
        E $\leftarrow$ ConcatEmbedding(\{$e_1$, $e_2$,...$e_{K-1}$, $e_r$\}) \\
        emb $\leftarrow$ ComputeEmbedding(X, $\theta$, E) \\
        \{$C_1$, $C_2$,...,$C_K$\} $\leftarrow$ ClusterData(emb, K) \\
        \{$e_1$, $e_2$,...$e_{K-1}$, $e_r$\} $\leftarrow$ SplitEmbedding(E, K) \\
        RC $\leftarrow$ False \\
    }

    \vspace{1mm}
    \Repeat(\Comment{Train all learners}){epoch completed}{
        $C_k \sim$ \{$C_1$, $C_2$,...,$C_K$\} \\
        b $\leftarrow$ GetBatch($C_k$) \\
        $L_k$ $\leftarrow$ FPass(b, $\theta$, $f^k$) \\
        $\theta$, $f^k$ $\leftarrow$ BPass($L_k$, $\theta$, $f^k$) \\
    }

    \vspace{1mm}
    ep $\leftarrow$ ep + 1 \\
    E $\leftarrow$ ConcatEmbedding(\{$e_1$, $e_2$,...$e_{K-1}$, $e_r$\}) \\
    RC $\leftarrow$ (ep mod $T_c$ == 0) \\
    
    \vspace{1mm}
    \If(\Comment{Is best}){Evaluate($X_{test}$, $\theta$, E, ep) $>$ B}{
        B $\leftarrow$ ep \\
    }
    \ElseIf(\Comment{Is network plateaued}){$ep \geq (B + T_p)$} {
    K $\leftarrow$ K + 1 \Comment{Update new learner} \\
    \{$e_{K-1}, e_r$\}  $\leftarrow$ splitLearner(\{$e_r$\}) \Comment{using Eq.\ref{eq:score}} \\
    \{$e_1$,..$e_{K-1}$, $e_r$\} $\leftarrow$ SplitEmbedding(E, K, $e_{K-1}$) \\
    reset($e_r$) \\
    RC $\leftarrow$ True \\
    }
}

\vspace{1mm}
E $\leftarrow$ ConcatEmbedding(\{$e_1$, $e_2$,...$e_{K-1}$, $e_r$\}) \\
$\theta$, E $\leftarrow$ FineTune(X, $\theta$, E) \\
\textbf{Output:}  $\theta$, E \\
\caption{Dynamic Subspace Learner Pseudocode}
\label{algo:DynK}
\end{algorithm}


\subsection{Attentive Dynamic Subspace Learners}
\label{sec:adsl}
Deep attention is raising as an efficient mechanism to focus the learning on the objects of interest in a wide range of applications, such as person re-identification \cite{li2018harmonious}, object classification \cite{wang2017residual}, or medical image segmentation \cite{sinha2019multi,schlemper2019attention}. Inspired by these advances, we introduce an attention module to learn attentive features, with the goal of enhancing the learning of the embedding space.
For a given input image $x_i$, feature extractor $S(\cdot)$ produces a feature maps $s_i=S(x_i) \in \mathbb{R}^{c \times m \times n}$, where $m,n$ denote the spatial dimension of the feature map and $c$ the number of channels. The attention map produced by the attention module $A(\cdot)$ can be then defined as $a_i = A(s_i) \in \mathbb{R}^{m \times n}$. The generated attention map is multiplied with each feature map $a_i \odot s_i$, where $\odot$ is the element-wise product, resulting in the set of attentive features. Last, the attentive features are combined to produce a $c-$dimensional vector by using global average pooling (GAP), which are mapped into the manifold space using a dense layer (\figref{fig:arch}).


\subsection{Attention maps for Weakly Supervised Segmentation}

The attention maps obtained by our proposed method can serve as proxy \textit{pixel-level} labels to train a segmentation network in a fully-supervised manner. Specifically, the input image $x_i$ and corresponding attention map $a_i$ are used as a training pair. To differentiate foreground pixels from the background pixels in $a_i$, we threshold the attention maps with $T_s$ (i.e., pixels in $a_i$ greater than $T_s$ are set to 1, 0 otherwise) before training the segmentation network. The network is trained with binary cross-entropy as a loss function, which is computed over pixel-wise softmax probabilities, defined as:


\begin{equation}
\mathcal{L}_{BCE}(x, a) = - \frac{1}{N}\sum_{i=1}^{N} \sum_{c=1}^{2} a_i^c \cdot  log(F_{\theta_s}(x_i^c)) 
\end{equation}

%
where $F_{\theta_s}$ is a segmentation network parameterized by $\theta_s$. Note that the learning objective that trains a segmentation network is same in both the fully and weakly supervised scenario. However, the main difference lies in the labels employed in the cross-entropy term. In particular, while the former resorts to given segmentation masks, e.g., $\mathbf{y}$, the latter leverages the obtained attention masks as pseudo-labels, i.e., $\mathbf{a}$.


\begin{figure*}[t!]
\centering
\begin{subfigure}{0.328\textwidth}
\centering
\includegraphics[width=1\linewidth,trim={1cm 0.2cm 1.5cm 1.2cm},clip]{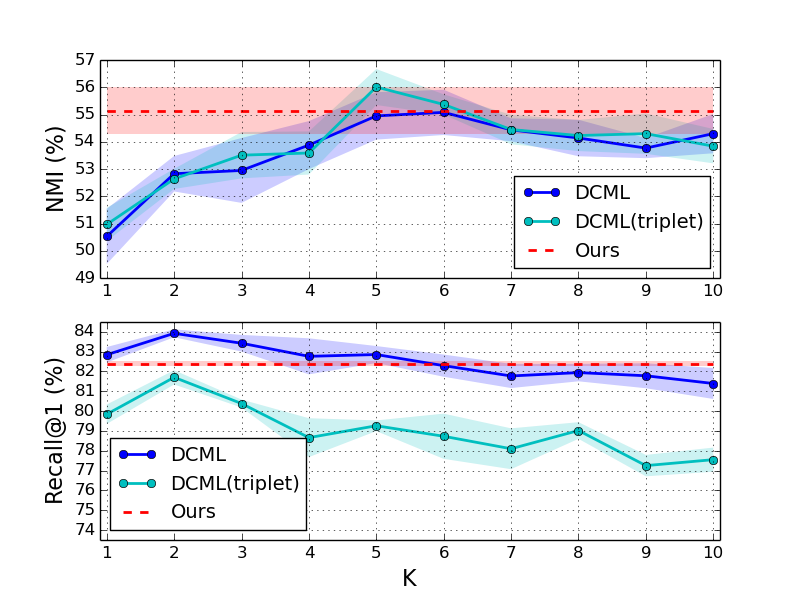}
    \caption{ISIC19 dataset}
    \label{fig:1}
\end{subfigure}
\begin{subfigure}{0.328\textwidth}
\centering
\includegraphics[width=1\linewidth,trim={1cm 0.2cm 1.5cm 1.2cm},clip]{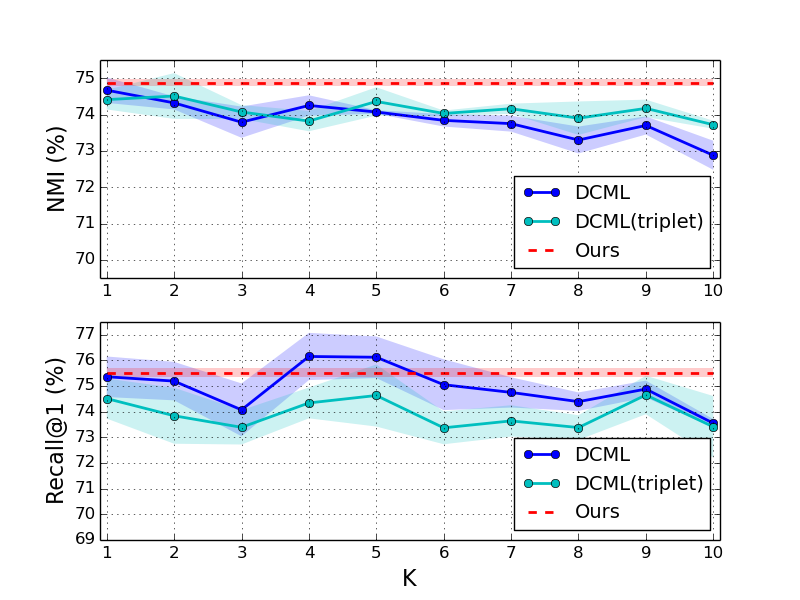}
    \caption{MURA dataset}
    \label{fig:2}
\end{subfigure}
\begin{subfigure}{0.328\textwidth}
\centering
\includegraphics[width=1\linewidth,trim={1cm 0.2cm 1.5cm 1.2cm},clip]{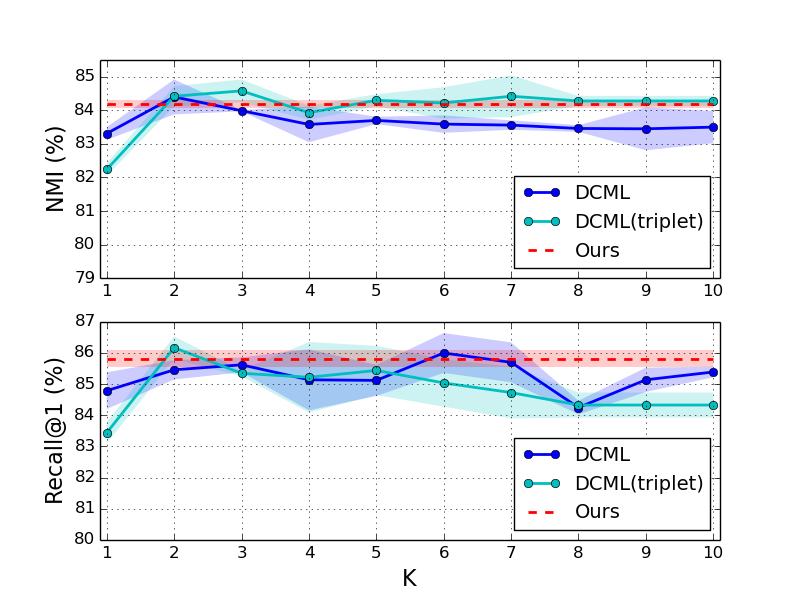}
    \caption{HyperKvasir dataset}
    \label{fig:3}
\end{subfigure}
\caption{\textbf{Impact of number of learners $K$ in DCML \cite{sanakoyeu2019divide}} - Each line indicates the NMI (top) and Recall@1 (bottom) scores across the three datasets. The default loss function employed is margin loss, whereas models with a triplet loss are explicitly mentioned. Best seen in color.}
\label{fig:impactLearner}
\end{figure*}



\section{Experiments}
\subsection{Experimental Setting}
The performance of the proposed attention-based dynamic subspace learners (ADSL) is compared to other deep metric learning methods applied in medical imaging \cite{teh2019metric,pati2020reducing,sikaroudi2020supervision,gupta2019deep,yan2018deep}, which resort to contrastive or triplet loss. To assess the effectiveness of the dynamic learner training strategy, we compare it with the divide and conquer approach (DCML) \cite{sanakoyeu2019divide}. Since we use class labels information, we compare with the classification network trained using a cross-entropy loss.
For a fair evaluation, the backbone architecture and hyper-parameters are fixed across the different methods. In addition, experiments across all the models and datasets are run three times, and their average performances are reported. Note that the baselines based on triplet and contrastive loss rely on single-learner, whereas models based on the \textit{divide-and-conquer} strategy and our method employ multiple learners.

To assess the performance of our approach in terms of segmentation, we benchmark the resulting attention maps against the popular GradCAM \cite{selvaraju2017grad} from the classification networks. 
We include a recent Attention Residual Learning (ARL) approach in \cite{zhang2019attention} since it has been similarly proposed in the context of skin lesion analysis. We also include a recently proposed weakly supervised segmentation method, Embedded Discriminative Attention Mechanism (EDAM) \cite{wu2021embedded}, applied for the natural image.
Lastly, we include as an upper bound the results obtained by UNet \cite{ronneberger2015u} that was trained on the provided pixel-level masks. Note that the model architecture and hyperparameters are fixed across the different methods. Nevertheless, the ARL model employs a carefully modified ResNet50 backbone with soft-attention blocks in each layer. It is noteworthy to mention that it also uses an offline multi-scale patch extraction strategy, resulting in extra images during training. Whereas, the EDAM model employs a collaborative multi-head attention module after the feature extraction layer to directly generate the discriminative activation masks.



\subsubsection*{Datasets}
The performance of the proposed method, in terms of clustering and image retrieval, is evaluated on three diverse medical imaging datasets: skin lesion from the ISIC 2019 Challenge \cite{codella2019skin,combalia2019bcn20000}, musculoskeletal radiographs from the MURA dataset \cite{rajpurkar2017mura}, and gastrointestinal tract images from the HyperKvasir dataset \cite{borgli2020hyperkvasir}. To assess the segmentation performance, we resort to the skin lesion dataset from the ISIC 2018 Challenge \cite{tschandl2018ham10000,codella2019skin}.


\paragraph{ISIC19} This dataset consists of 25,331 images across 8 different categories. In our experiments, following the standard procedure in DML, we split our dataset into independent training and testing sets. Specifically, 20,000 images were used for training and the remaining 5,331 for testing.

\paragraph{MURA} It consists of 40,561 images from 9,045 normal and 5,818 abnormal musculoskeletal radiography studies across seven standard upper extremity types. We configure this as 14 categories (7 normal and 7 abnormal) to represent the data in a manifold. We use the provided split of 36,808 images for training and 3,197 images for testing.

\paragraph{HyperKvasir} This dataset consists of 110,079 images, of which 10,662 images are labeled across 23 different classes of findings. We randomly split the data into 8,567 images for training and the remaining 2,095 images for testing.

\paragraph{ISIC18}
This dataset is composed of 2,594 images and their corresponding pixel-level masks. The segmentation dataset is randomly split into three sets: training (1,042), validation (520), and testing (1,038). We leverage the attention maps and GradCAMs generated on the ISIC19 dataset (25,331 images) as proxy-labels to train the segmentation networks. In contrast, the training set is used to train the upper-bound model, i.e., fully-supervised.

\subsubsection*{Evaluation Metrics}
We follow the evaluation protocol typically employed in deep metric learning \cite{sanakoyeu2019divide,oh2016deep}. In particular, we employ the normalized mutual information (NMI) to assess the clustering performance using K-means and the Recall score (with k = 1 and 4) to evaluate the image retrieval quality. To assess the segmentation performance, we employ the common Dice score coefficient.


\subsubsection*{Implementation details}
\label{sec:sml_Impn}
As in \cite{sanakoyeu2019divide}, we use ResNet50 \cite{he2016deep} as the backbone architecture. The feature extractor layers consist of the first three residual blocks of ResNet50, used as input to the attention module.
The attention module consists of three convolution layers with $3 \times 3$ kernel and filters size of \{128, 32, 1\}, with a ReLU activation between each convolutional layer. Last, a sigmoid activation is integrated into the final layer to produce the activation map. An input image size of $224 \times 224$ is used for all our experiments. All models are trained using the Adam optimizer \cite{kingma2014adam} with batch size of $B$ = 32. In each mini-batch, 8 images per class are sampled to ensure a class-balanced scenario and experiments are trained for 300 epochs. The last 50 epochs are fine-tuned with full embedding. The re-clustering parameter is set to $T_c$ = 2 as in \cite{sanakoyeu2019divide} and the network plateau threshold is empirically set to $T_p$ = 10. The margin loss parameters are set to $\alpha$ = 0.2, $\beta$ = 1.2, as in \cite{wu2017sampling}. Last, since most DML approaches \cite{wu2017sampling,sanakoyeu2019divide} employ an embedding space of size $d$ = 128, we use the same latent dimension in all our experiments.

Regarding the segmentation task, we use UNet \cite{ronneberger2015u} architecture with an initial kernel size of 32 with two convolution layers and a depth of 3. It is trained with Adam optimizer with batch sizes of 16. For each method, the threshold parameter $T_s$ is set to maximize the Dice score on the initial maps of the validation set (\figref{fig:segThreshold}).

\subsection{Clustering and image retrieval results}
\subsubsection*{Impact of number of learners $K$}
One of the motivations of this work is to remove the need to empirically searching for the optimal number of learners. To validate this hypothesis, we first study the performance of DCML \cite{sanakoyeu2019divide} by varying the number of subspace learners ($K$). Figure~\ref{fig:impactLearner} depicts the results of this experiment across the three datasets and under two different loss functions: margin and triplet loss. In these plots, it can be observed that the optimal $K$ value significantly differs across datasets and metrics. Thus, this limitation of the DCML approach results in extra time-consuming steps to fine-tune the model in each dataset. In contrast, the proposed method (dotted line) eliminates the need of manually defining $K$ by dynamically exploring the manifold, yet achieves on par results with the best performing DCML setting.

We also report the average $K$ values obtained from our method over three runs, as well as the DCML (best) $K$ in Table~\ref{table:obtainedK}. The table shows that the $K$ value has no relation to the number of ground-truth classes. The dynamically obtained $K$ in our method is driven by image content, not by the number of ground-truth classes, which explains their uncorrelated values.

\begin{table}[ht!]
\centering
\addtolength{\tabcolsep}{3pt}
\begin{tabular}{l | c | c | c }
\bf Dataset    & \bf \#classes & \bf ADSL - Avg. K   & \bf DCML - Best K \\ [0.2ex] 
\midrule
ISIC19  & 8 & 7 & 6 \\
MURA  & 14 & 4.67 & 1 \\
HyperKvasir & 23 & 4.33 & 2 \\
\end{tabular}
\caption{Comparison of the obtained $K$ values from our method and the DCML best K values with respect to the number of ground-truth classes.}
\label{table:obtainedK}
\end{table}



\begin{table*}[t!]
\centering
\addtolength{\tabcolsep}{14.0pt}
\begin{tabular}{l | c  c  c | c}
Method              & \bf NMI ($\uparrow$)           & \bf R@1 ($\uparrow$)           & \bf R@4 ($\uparrow$) & \bf Avg. of NMI + R@1 ($\uparrow$)\\ [0.2ex] 
\midrule
Classification network  & 45.41 $\pm$ 1.95   & 77.85 $\pm$ 0.86   & 90.54 $\pm$ 0.51   & 61.63 $\pm$ 1.40 \\
Contrastive loss  & 31.47 $\pm$ 0.39   & 78.13 $\pm$ 0.59   & 91.13 $\pm$ 0.08  & 54.80 $\pm$ 0.49 \\
Triplet loss      & 50.97 $\pm$ 0.61   & 79.84 $\pm$ 0.49   & 91.70 $\pm$ 0.26  & 65.41 $\pm$ 0.55 \\

DCML (worst NMI, K = 1)      & 50.53 $\pm$ 1.01   & \bf 82.84 $\pm$ 0.39   & 91.51 $\pm$ 0.43  & 66.69 $\pm$ 0.70 \\
DCML (best NMI, K = 6)       &  \ul{55.08 $\pm$ 0.83}   & 82.29 $\pm$ 0.56   & \ul{91.73 $\pm$ 0.36}  & \ul{68.69 $\pm$ 0.70} \\

\rowcolor{gray!20} ADSL (free from K, ours)       & \bf 55.14 $\pm$ 0.87   & \ul{82.39 $\pm$ 0.11}   & \bf 92.11 $\pm$ 0.27 & \bf 68.77 $\pm$ 0.49 \\ 

\end{tabular}
\caption{\textbf{Quantitative evaluation on ISIC19 test set} - The NMI, Recall, and average scores from the different methods. Our method is emphasized with light gray, whereas best and second-best results are highlighted with bold and underline.}
\label{table:scoreISIC}
\end{table*}

\begin{table*}[t!]
\centering
\addtolength{\tabcolsep}{14.0pt}
\begin{tabular}{l | c  c  c | c }
Method              & \bf NMI ($\uparrow$)           & \bf R@1 ($\uparrow$)           & \bf R@4 ($\uparrow$)  & \bf Avg. of NMI + R@1 ($\uparrow$) \\ [0.2ex] 
\midrule
Classification network  & 71.09 $\pm$ 1.25   & 74.21 $\pm$ 0.27   & 92.59 $\pm$ 0.40  & 72.65 $\pm$ 0.76  \\
Contrastive loss    & 74.28 $\pm$ 0.53  & 71.65 $\pm$ 0.53  & 92.07 $\pm$ 0.36    & 72.97 $\pm$ 0.53  \\
Triplet loss        & 74.41 $\pm$ 0.27  & 74.51 $\pm$ 0.78  & \bf 92.95 $\pm$ 0.33    & 74.46 $\pm$ 0.53  \\

DCML (worst NMI, K = 10)   & 72.88 $\pm$ 0.40   & 73.55 $\pm$ 0.16    & 91.17 $\pm$ 0.19      & 73.22 $\pm$ 0.28  \\
DCML (best NMI, K = 1)    & \ul{74.67 $\pm$ 0.35}   & \ul{75.36 $\pm$ 0.79}    & \ul{92.89 $\pm$ 0.18}      & \ul{75.02 $\pm$ 0.57}  \\

\rowcolor{gray!20} ADSL (free from K, ours)     & \bf 74.88 $\pm$ 0.09   & \bf 75.52 $\pm$ 0.18  & 92.25 $\pm$ 0.42    & \bf 75.20 $\pm$ 0.15  \\

\end{tabular}
\caption{\textbf{Quantitative evaluation on MURA test set} - The NMI, Recall, and average scores from the different methods. Our method is emphasized with light gray, whereas best and second-best results are highlighted with bold and underline.}
\label{table:scoreMURA}
\end{table*}

\begin{table*}[t!]
\centering
\addtolength{\tabcolsep}{14.0pt}
\begin{tabular}{l | c  c  c | c }
Method              & \bf NMI ($\uparrow$)           & \bf R@1 ($\uparrow$)           & \bf R@4 ($\uparrow$) & \bf Avg. of NMI + R@1 ($\uparrow$) \\ [0.2ex]  
\midrule
Classification network  & 80.13 $\pm$ 2.34   & \ul{85.66 $\pm$ 0.39}   & \bf 94.42 $\pm$ 0.39   & 82.90 $\pm$ 1.87   \\
Contrastive loss    & 83.89 $\pm$ 0.15  & 78.52 $\pm$ 0.86  & 93.44 $\pm$ 0.48  & 81.21 $\pm$ 0.51   \\
Triplet loss        & 82.24 $\pm$ 0.19  & 83.44 $\pm$ 0.34  & 93.92 $\pm$ 0.22  & 82.84 $\pm$ 0.27   \\

DCML (worst NMI, K = 1)    & 83.31 $\pm$ 0.19   & 84.79 $\pm$ 0.59    & 94.05 $\pm$ 0.26    & 84.05 $\pm$ 0.39   \\
DCML (best NMI, K = 2)    & \bf{84.40 $\pm$ 0.52}   & 85.46 $\pm$ 0.31    & 94.19 $\pm$ 0.28    & \ul{84.93 $\pm$ 0.42}   \\

\rowcolor{gray!20} ADSL (free from K, ours)      & \ul{84.18 $\pm$ 0.12}   & \bf 85.82 $\pm$ 0.27  & \ul{94.24 $\pm$ 0.41}  & \bf 85.00 $\pm$ 0.20   \\ 

\end{tabular}
\caption{\textbf{Quantitative evaluation on HyperKvasir test set} - The NMI, Recall, and average scores from the different methods. Our method is emphasized with light gray, whereas best and second-best results are highlighted with bold and underline.}
\label{table:scoreHK}
\end{table*}





\subsubsection*{Comparison to prior literature}
We now compare our method with recent prior work as baselines, whose results are reported in Tables~\ref{table:scoreISIC}-\ref{table:scoreHK}. As the performance of DCML varies with $K$, we report only the best and worst models. Note that the DCML with a single-learner, i.e., $K=1$, is equivalent to a margin loss method \cite{wu2017sampling}. We also report the performance of the embedding space learned by the classification network. From the Tables~\ref{table:scoreISIC}-\ref{table:scoreHK}, we observe that the proposed method consistently achieves the best results in terms of NMI across the three datasets while performing on par with the best setting of the DCML approach on image retrieval metrics. As shown previously, it is important to note that the performance of DCML heavily depends on the value of $K$. For instance, the difference between the worst and best DCML configuration in NMI score can be up to 5\% on the ISIC19 dataset. Compared to single-learner approaches, our method brings 5 and 2\% improvements in NMI and Recall score on the ISIC19 dataset and up to a 1\% improvement in both scores on the MURA and HyperKvasir datasets. This highlights the potential of exploring embeddings via multiple subspaces. 

Furthermore, the comparison with the conventional classification network shows that our method consistently outperforms its accuracy up to 10\% in terms of NMI scores on ISIC19 and up to 4\% NMI score on MURA and HyperKvasir datasets and up to 4\% and 1.5\% in terms of Recall scores on the ISIC19 and MURA datasets. The averaged NMI and R@1 results of the proposed method slightly outperform the best DCML configuration, which is consistent across all the datasets. The standard deviation of our method is smaller in all cases for all metrics compared to the DCML. Overall, our method shows better robustness with respect to the state-of-the-art methods in the learning manifold space. The performance of our method is in line with the recent literature \cite{barata2021improving,allegretti2021supporting}.

\begin{table}[t!]
\centering
\addtolength{\tabcolsep}{3.5pt}
\begin{tabular}{l | c | c  c  c }
Dataset             & Method               & \bf NMI ($\uparrow$)           & \bf R@1 ($\uparrow$)           & \bf R@4 ($\uparrow$) \\ [0.5ex] 
\toprule

\multirow{2}{*}{ISIC19}
& DSL    & 54.11 &  \bf 82.74  & 91.95  \\ 
& ADSL    & \bf 55.14 & 82.39 & \bf 92.11 \\ 
\midrule

\multirow{2}{*}{MURA} 
& DSL    & 74.21 & \bf 75.85 & \bf 92.26 \\ 
& ADSL    & \bf 74.88 & 75.52 & 92.25 \\ 
\midrule

\multirow{2}{*}{HyperKvasir}
& DSL    & \bf 84.44 & 85.36 & 93.54 \\ 
& ADSL    & 84.18 & \bf 85.82 & \bf 94.24 \\  
[0.5ex] 
\bottomrule
\multirow{2}{*}{\bf Average}
& DSL    &  70.92 & \bf 81.32 &  92.58 \\ 
& ADSL    & \bf 71.40 &  81.24 & \bf 92.87 \\  
\end{tabular}
\caption{\textbf{Impact of attention module} - Per-dataset and average results of the proposed model with (ADSL) and without (DSL) the attention module. Best results are highlighted with bold for each dataset as well as for the average.}
\label{table:scoreAttention}
\end{table}

\begin{figure}[h!]
\centering
\includegraphics[width=0.915\linewidth]{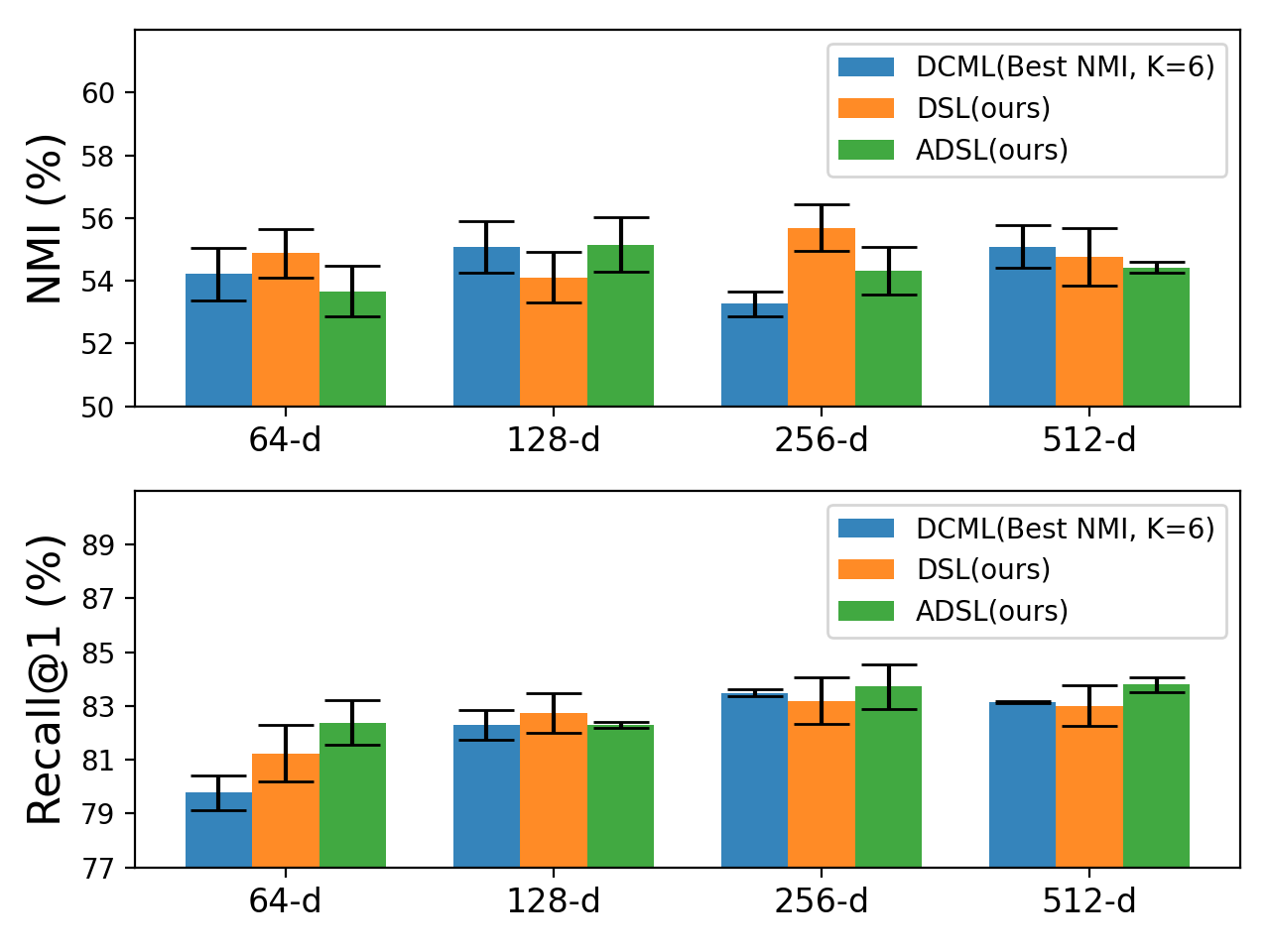}
\caption{\textbf{Impact of the embedding size} - Each bar indicates the NMI (top) and Recall@1 (bottom) scores on ISIC19 dataset. Compare to the best model of DCML method, our method produces better NMI and Recall scores for most cases.}
\label{fig:emb_dim}
\end{figure}

\subsubsection*{Ablation study on the use of attention}
Adding an attention module brings additional value to our model in terms of interpretability. Nevertheless, to assess whether this improvement is also reflected in the model performance, we compare our model to its non-attention counterpart, denoted as Dynamic Subspace Learners (DSL). Results from this study are reported in Table \ref{table:scoreAttention}, which shows that adding attention typically leads to a boost on the model performance. In particular, the attentive model brings 0.5 and 0.3\% improvement as average over the three datasets for the NMI and R@4 metrics, respectively, while achieves on par results for R@1. Additionally, the attention module minimally increases the model memory by 5 MB (includes parameters, forward and backward pass size) when compared to non-attention counterpart, which is arguably negligible with respect to the overall model size (607 MB) in case of deployment.


\subsubsection*{Impact of the embedding size}
We also evaluate the effect of representing the embedding space with different sizes. In particular, we assess the clustering and image retrieval performance on the ISIC19 dataset by fixing the embedding dimension size to 64, 128, 256, and 512. Figure~\ref{fig:emb_dim} shows that increasing the embedding size results in a performance improvement, which is reflected in both NMI and recall metrics. Nevertheless, beyond a 256-dimension embedding, the performance of both models typically decreases. 


\begin{figure*}[h!]
\centering
\begin{subfigure}{0.195\textwidth}
\centering
\includegraphics[width=1\linewidth,trim={1cm 0.2cm 1.5cm 1.2cm},clip]{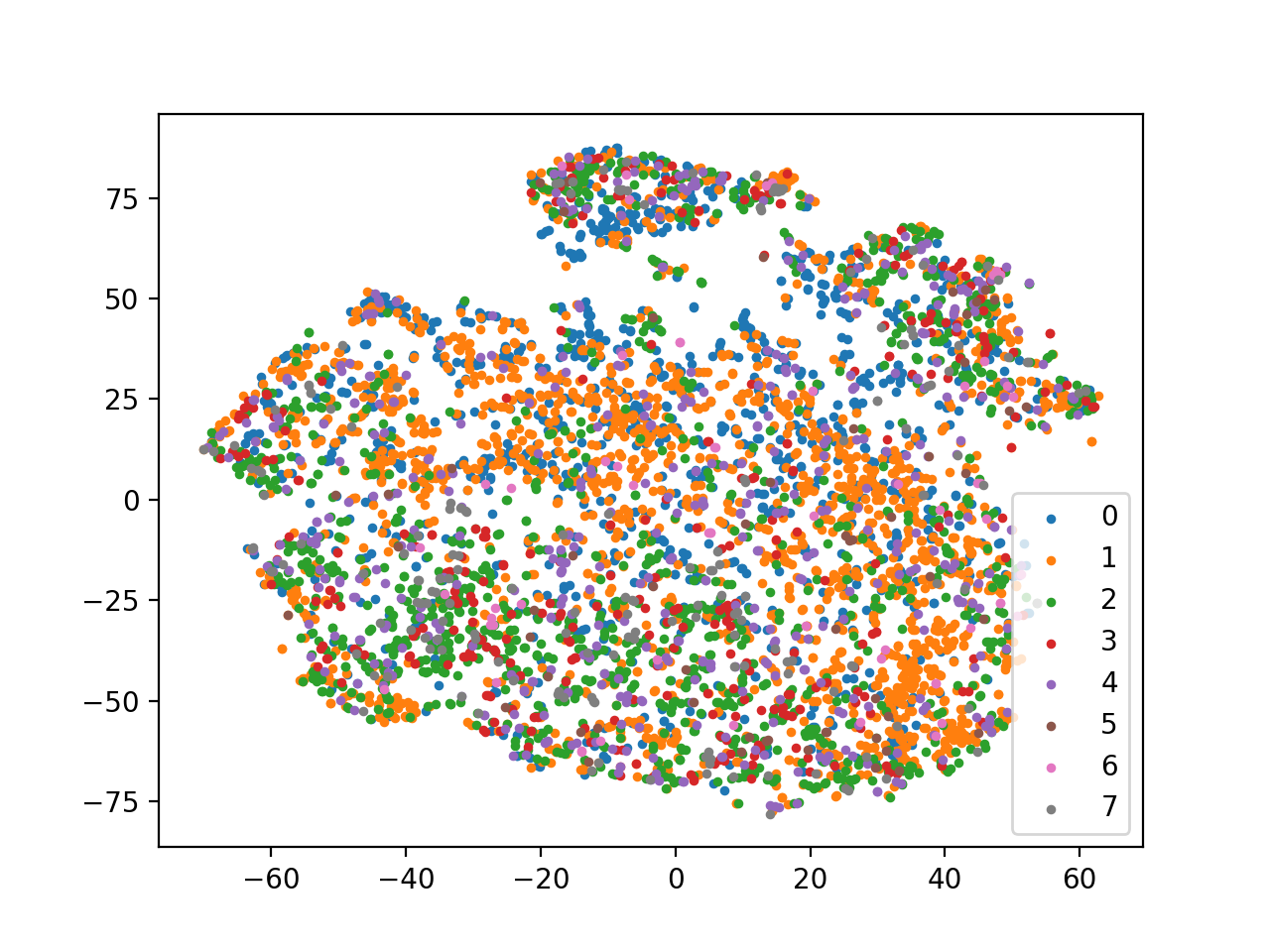}
    \caption{Before learning}
    \label{fig:emb_before}
\end{subfigure}
\begin{subfigure}{0.195\textwidth}
\centering
\includegraphics[width=1\linewidth,trim={1cm 0.2cm 1.5cm 1.2cm},clip]{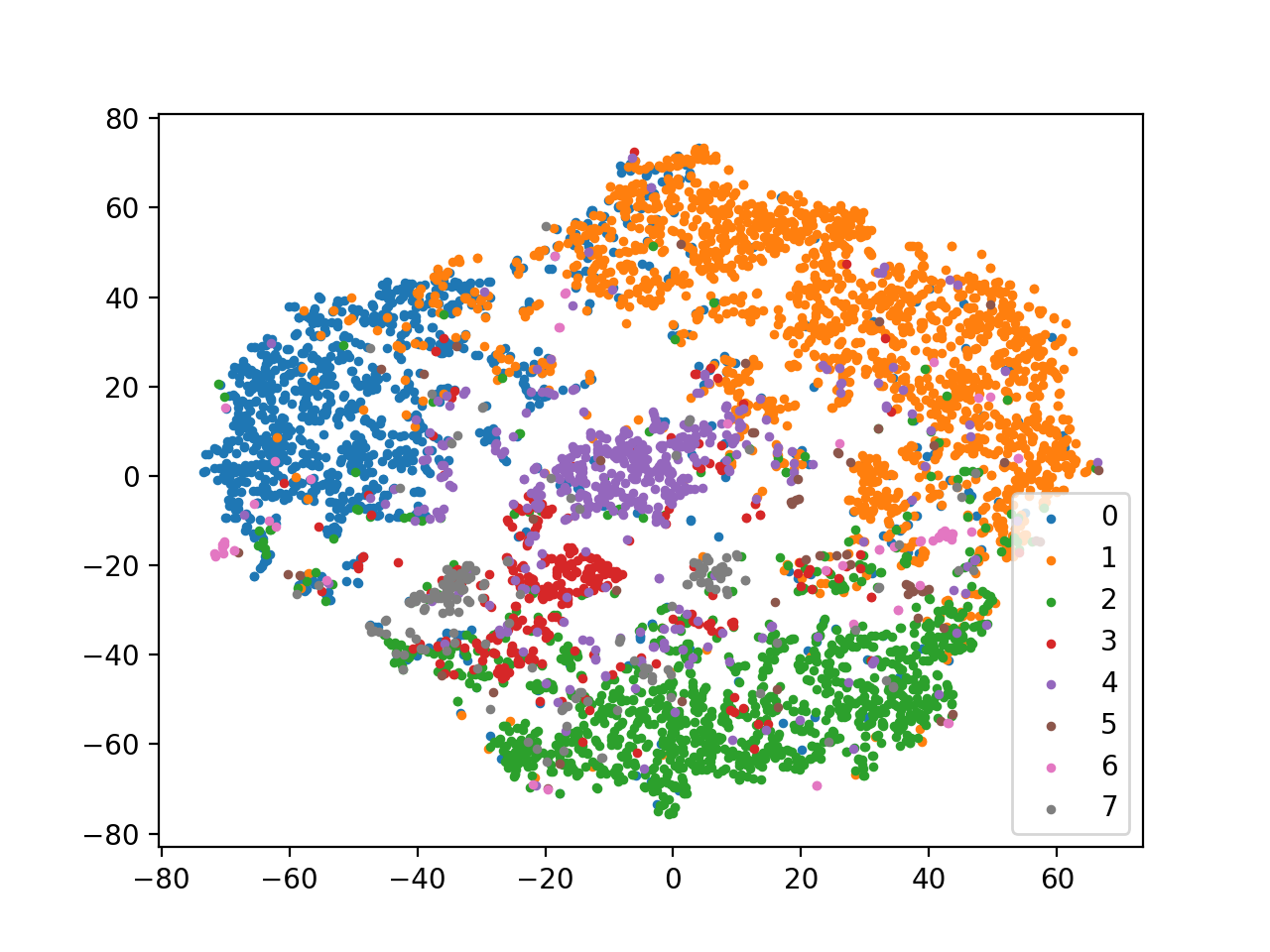}
    \caption{Classification network}
    \label{fig:emb_cls}
\end{subfigure}
\begin{subfigure}{0.195\textwidth}
\centering
\includegraphics[width=1\linewidth,trim={1cm 0.2cm 1.5cm 1.2cm},clip]{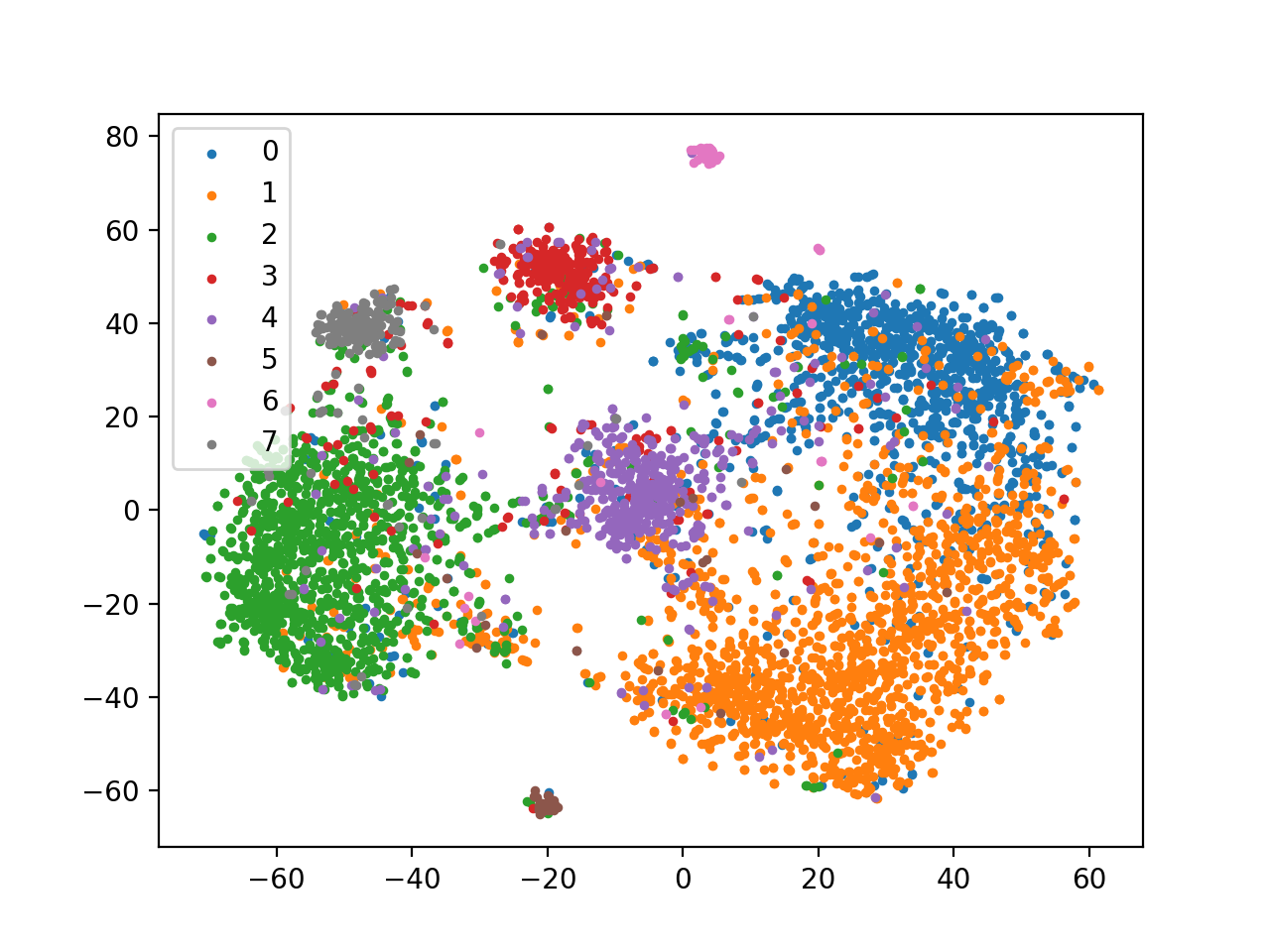}
    \caption{DCML $K=1$}
    \label{fig:emb_dcml1}
\end{subfigure}
\begin{subfigure}{0.195\textwidth}
\centering
\includegraphics[width=1\linewidth,trim={1cm 0.2cm 1.5cm 1.2cm},clip]{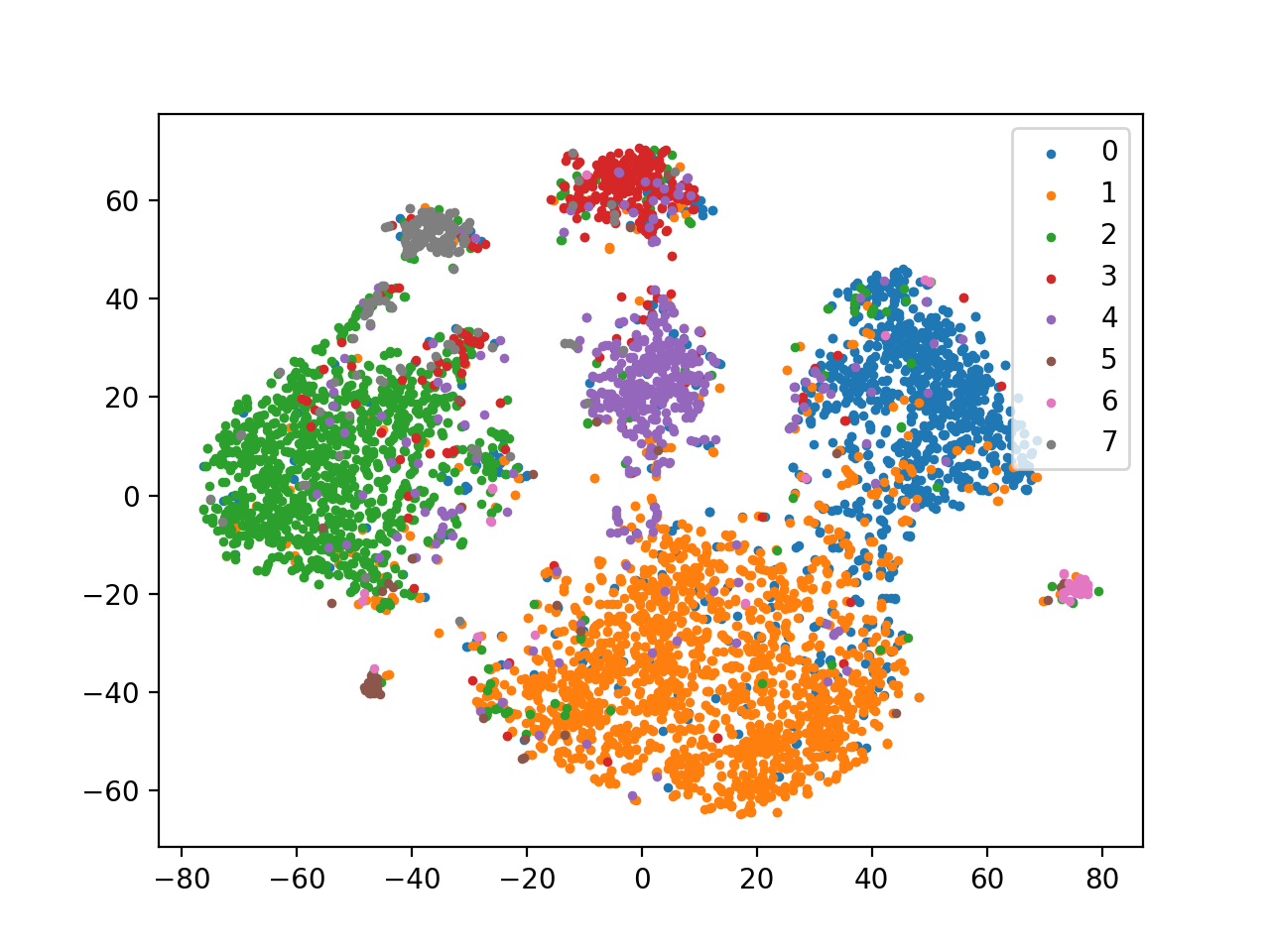}
    \caption{DCML $K=6$}
    \label{fig:emb_dcml6}
\end{subfigure}
\begin{subfigure}{0.195\textwidth}
\centering
\includegraphics[width=1\linewidth,trim={1cm 0.2cm 1.5cm 1.2cm},clip]{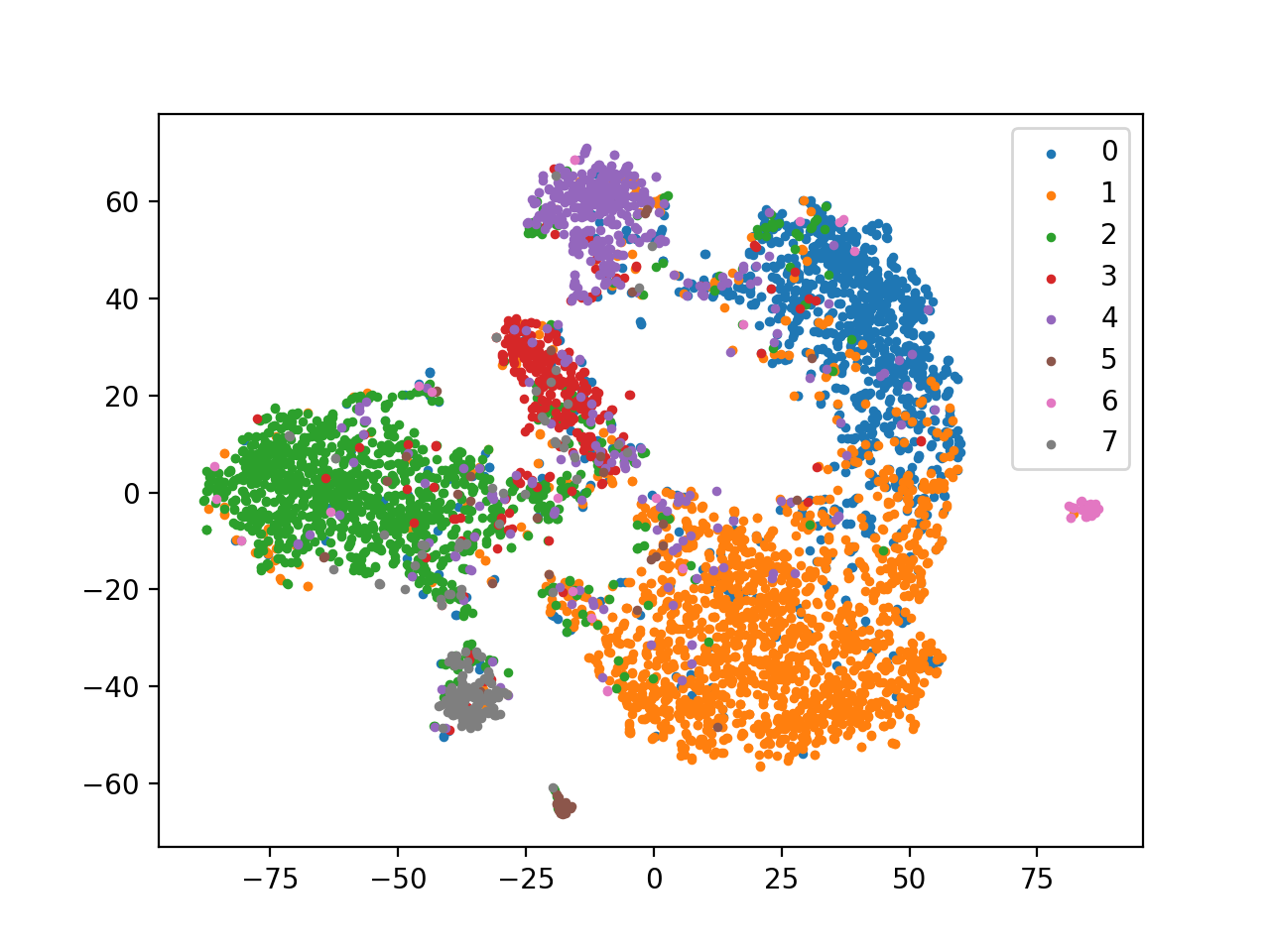}
    \caption{Our method}
    \label{fig:emb_adsl}
\end{subfigure}
\caption{\textbf{Visualization of ISIC19 test set in embedding space using t-SNE} - Each class is indicated by its individual color. When compared to a standard classification network, DCML $K=1$ (a single-learner) improves the separation between classes. The multi-learner methods, DCML $K = 6$ and our method, further improve the separation between classes, while our method has the advantage of being free from the number of learners $K$. Best seen in color.}
\label{fig:embedding}
\end{figure*}




\begin{figure}[t!]
\centering
\includegraphics[width=0.985\linewidth]{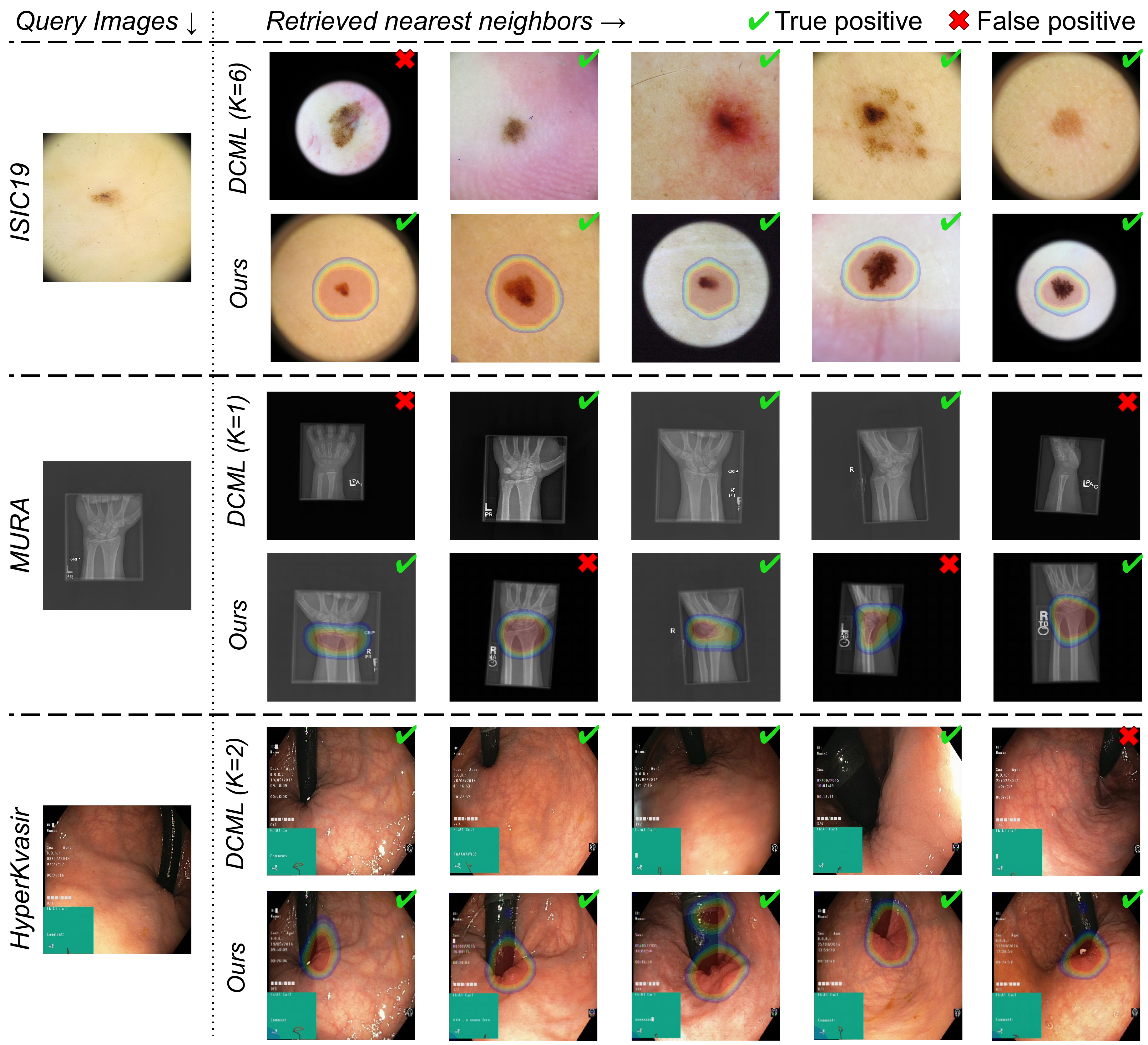}
\caption{\textbf{Performance of image retrieval on test sets} - Each query image and its five nearest neighbors in ascending order of distance are shown (left to right) from the DCML (best K) and our method with an overlay of our attention maps (probability above 0.5).}
\label{fig:retrievalADSLvsDCML}
\end{figure}


\subsubsection*{Qualitative Analysis}
To show the inter and intra-class representation power in the embedding space across different models, we visualize a t-SNE mapping \cite{maaten2008visualizing} on the ISIC19 test set (\figref{fig:embedding}). The classification network fails to discover clear boundaries across classes in the embedding space (\figref{fig:embedding}b). This could be because of the cross-entropy loss when coupled with softmax, does not explicitly guarantee the minimization of intra-class variance or maximization of inter-class variance, which results in suboptimal discriminative features \cite{liu2016large}.
The single metric learner, i.e., DCML $K = 1$ (\figref{fig:embedding}c), improves the class boundaries when compared to the classification network, yet they fail to possess compact clusters. On the other hand, inter-class discrimination is visually enhanced when resorting to multiple learners, i.e., DCML $K = 6$ (\figref{fig:embedding}d) and our approach (\figref{fig:embedding}e). Further, we can also observe that the proposed model yields more compact clusters than the DCML approach, which might be due to the freedom of our model to explore the manifold.


Qualitative evaluation in terms of image retrieval is assessed in \figref{fig:retrievalADSLvsDCML}, where a given random query with its five nearest neighbors, found using both DCML and our method, are shown. Additionally, we overlay the contour of our attention maps (having probability above 0.5) from the proposed method over their respective retrieved image. First, our method indeed retrieves images having similar lesions and colors from the ISIC19 dataset. In radiography wrist images, both DCML and our method have similar retrieval errors. Finally, retrieval images from the HyperKvasir dataset have similar image semantics in terms of texture and probe length using our method when compared to DCML. The coherence of image retrievals indicates that the intra- and inter-class similarities have been captured by our method and thereby demonstrates the robustness of our learned embedding. Moreover, our attention maps mainly concentrate on the lesion in the skin images, the wrist in the radiography images, and the probe contact region in the endoscopic images, demonstrating that our model decision are consistent over all retrievals.

\subsection{Weakly Supervised Segmentation results}
Table ~\ref{table:dicescore} reports the results of the segmentation experiments. In this table, \textit{Init maps} are used to denote the raw visual salient regions from either GradCAM or attention maps. \textit{Refined} refers to the performance of the segmentation network trained on the \textit{Init maps}. First, we can observe that segmentation results obtained by raw attention maps and GradCAMs are considerably low, with Dice values around 40\%. This is likely due to the well-known fact that both are highly discriminative, resulting in over-segmented regions. The Attention Residual Learning (ARL) significantly outperforms these baselines, whose improvement could be due to the use of attentive residual blocks and additional multiscale data augmentation. The attention maps from the recent Embedded Discriminative Attention Mechanism (EDAM) method perform at a similar level when compared to ARL.
Last, the attention maps from the proposed approach bring a significant boost compared to all the other methods. In particular, our model outperforms the baselines by nearly 30\% and the recent ARL model by 13\%. These results are typically consistent if we employ the initial maps as proxy-labels to train a segmentation network. In this case, raw attention maps or GradCAMs barely improve or even decrease the initial segmentation performance. In contrast, ARL, EDAM, and the proposed method reach higher Dice values, with about 1\%, 3.5\%, and 3\% of increase, respectively. This represents a difference of 15\% in Dice with respect to ARL. On the other hand, by only using image-level information, the proposed model bridges the gap with a fully-supervised network, with only 14\% of difference. This suggests that the proposed model generates reliable segmentations.

\begin{table}[ht!]
\centering
\addtolength{\tabcolsep}{5pt}
\begin{tabular}{l | c | c }
\bf Method                      & \bf Init maps & \bf Refined \\ [0.5ex]
\toprule
Attention $^\ast$               & 38.45     & 33.43  \\
Attention $^\dagger$            & 38.52     & 38.38  \\
GradCAM $^\ast$                 & 41.55     & 40.76  \\
GradCAM $^\dagger$              & 39.80     & 41.27  \\
ARL \cite{zhang2019attention} $^\diamond$   & 56.78     & 57.60  \\
EDAM \cite{wu2021embedded} $^\diamond$      & 51.99 & 55.50 \\
ADSL (ours) $^\diamond$                   & \textbf{69.23} & \textbf{72.42}  \\
\midrule
Full-supervision (upperbound)   &  -  & 86.15  \\ [0.5ex]
\end{tabular}
\caption{\textbf{Performance of weakly supervised segmentation} - ``Initial maps'' and ``Refined'' are Dice scores (in \%) on the ISIC18 test set for different methods. Our method yields the best results (in bold). $^\ast$, $^\dagger$ and $^\diamond$ are from ResNet50, ResNet101 and modified ResNet50, respectively, indicating the used architecture in each visual map.}
\label{table:dicescore}
\end{table}


\subsubsection*{Ablation study of threshold $T_s$ on the raw visual maps}
\begin{figure}[t!]
\centering
\includegraphics[width=0.985\linewidth]{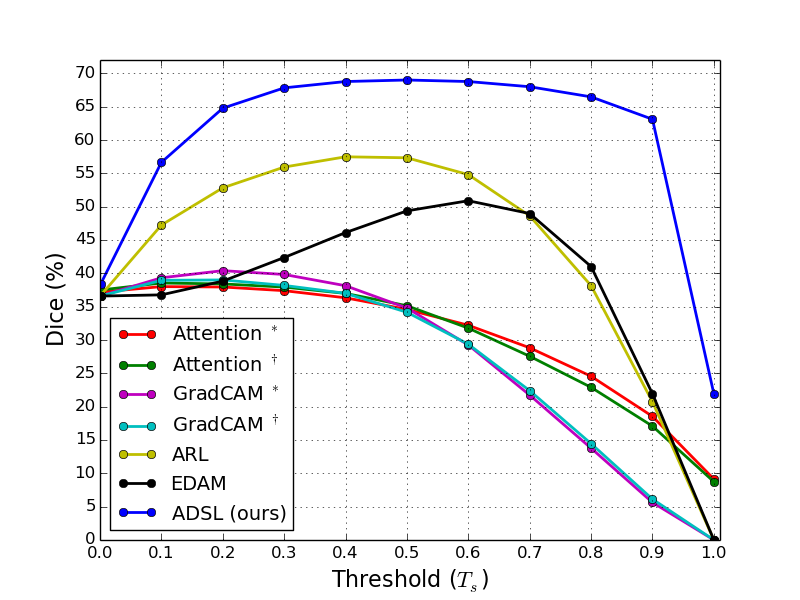}
\caption{\textbf{Threshold $T_s$ selection} - Each line indicates the Dice scores of initial maps on the ISIC18 validation set for different methods. Our method outperforms the baselines for all $T_s$ values. $^\ast$ and $^\dagger$ are obtained by classification networks using ResNet50 and ResNet101, respectively.}
\label{fig:segThreshold}
\end{figure}

We evaluate the effect of threshold values $T_s$ on the Dice score for raw visual maps from attention maps and GradCAMs, as shown in Fig~\ref{fig:segThreshold}. First, the attention maps and GradCAMs from the classification network have an almost flat Dice score of around 40\% until $T_s=0.4$, succeeded by a gradual decrease. The ARL and EDAM have a gradually increasing Dice score until $T_s=0.4$ and $T_s=0.6$ with a maximum score of 57.33\% and 50.89\%, respectively, followed by a gradual decrease. Our method outperforms the baselines for all threshold values in Dice scores with a maximum dice score of 69.0\%, showing the robustness of the attention maps derived from our method. This study assists in setting a threshold value $T_s$ for each method before training the segmentation network.

\begin{figure*}[t]
\centering
\includegraphics[width=0.985\linewidth]{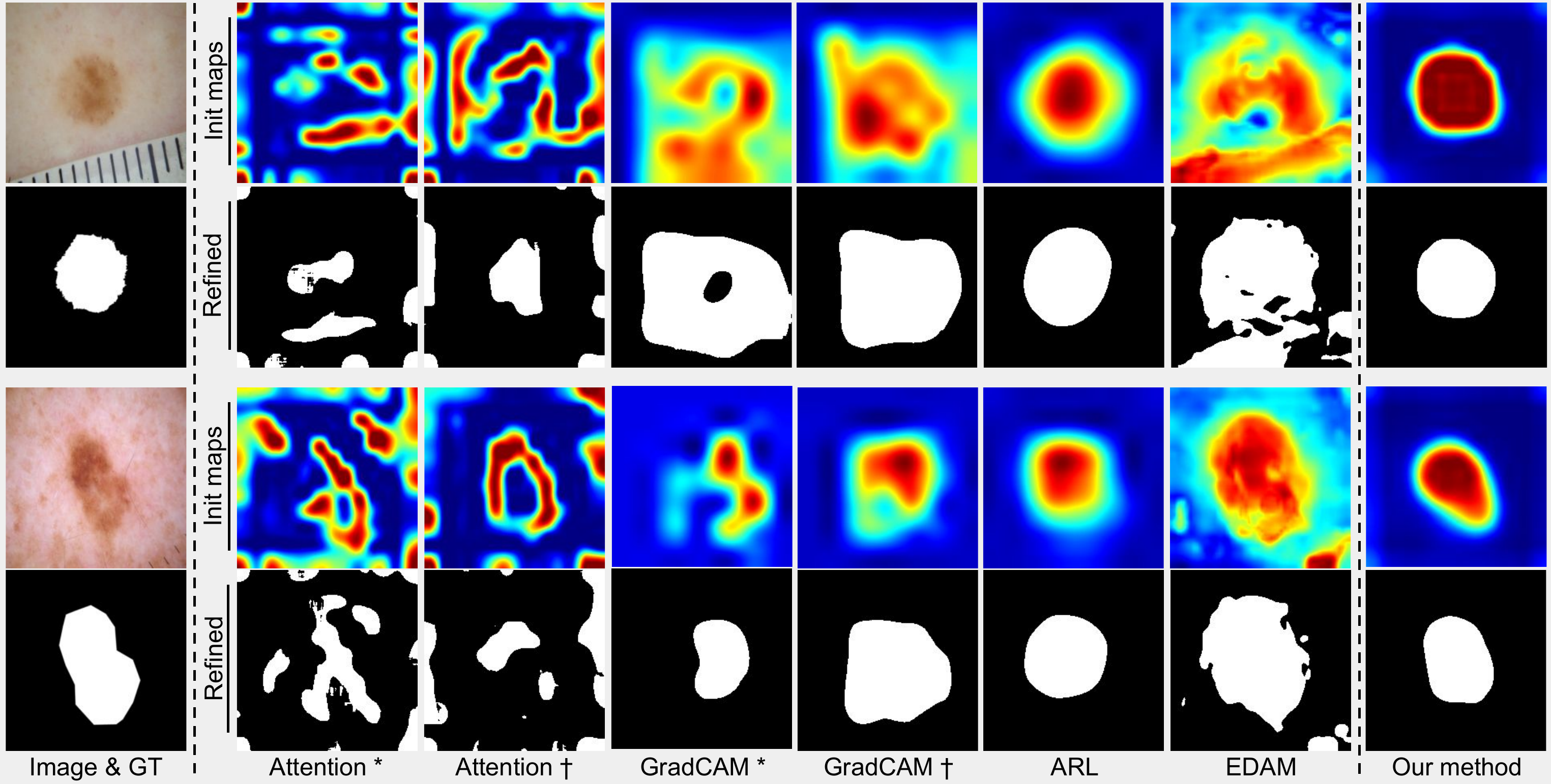}
\caption{\textbf{Visual results of segmentation} - \textit{(Init maps)} Saliency map obtained by different methods and \textit{(Refined)} their segmentation results. $^\ast$ and $^\dagger$ are obtained by GradCAM on classification networks using ResNet50 and ResNet101, respectively.}
\label{fig:seg}
\end{figure*}

\subsubsection*{Qualitative Performance Evaluation}
Visual results of the different methods are shown in Fig. ~\ref{fig:seg}. In this figure, \textit{Init maps} (row 1 and 3) are raw visual salient regions from either GradCAM or attention maps shown as heatmaps, whereas \textit{Refined} (row 2 and 4) refers to the performance of the segmentation network trained using \textit{Init maps} as a proxy-labels. The attention maps (row 1 and 3) produced by the classification network spread all over the image, capturing some discriminative regions on the target lesion. GradCAMs spread around the target, highlighting discriminative regions of the lesion but failing to capture the whole context. The saliency map produced by the ARL method is focused on the target lesion. The attention maps obtained by the recent EDAM method spread around the target lesion, including the artifact regions, and fail to capture the target object context. In contrast, the attention maps derived from our approach better capture the attentive region, which mostly cover the lesion regions. The results show that our proposed approach generates superior attention maps compared to attention maps or GradCAMs from classification networks.

The results obtained by training a segmentation network on the initial salient regions (row 1 and 3) are depicted in row 3 and 4. These images demonstrate the feasibility of our method to weakly generate pixel-level labels that are usable for training segmentation networks.



\section{Discussion and Conclusion}
This paper presents a novel attention-based dynamic subspace metric learning approach for medical image analysis. The proposed algorithm leverages recent advances in deep metric learning using multiple metric learners. Our contribution improves the state-of-the-art method \cite{sanakoyeu2019divide} with dynamic exploitation of subspace learners to learn the embedding space. Specifically, our novel training strategy overcomes the empirical search of the optimal number of subspace learners parameter while achieving competitive results in clustering and image retrieval tasks. Performance is extensively evaluated on three publicly available benchmark datasets: skin lesions, musculoskeletal radiography, and endoscopic images. Results demonstrate that our dynamic learner approach achieves the best results in clustering performance across all three datasets. Compared to the single-learner method, our method brings a maximum of 5 and 2\% improvements in clustering and image retrieval scores on the ISIC19 dataset. Furthermore, our method significantly outperforms the classification network in all the datasets with a maximum of 10\% and 4\% improvements in clustering and retrieval scores on the ISIC19 dataset. Overall, the proposed method slightly outperforms in averaged results and has a smaller standard deviation when compared to the state-of-the-art methods in multiple metric learning. Our experiments have shown consistency across all the datasets, demonstrating the robustness of our method. Qualitative results show that the proposed method produces compact clustering and coherence image retrievals.  

The addition of the attention module to our subspace learners provides the visual interpretability of the learned embedding space in terms of attention maps and improves the clustering metrics. Our method offers new tools in multiple metric learners approaches, notably dynamically learning the number of learners and providing attention maps to hint at salient information caught by the learners. Studying the clinical usability of these tools remains to be explored. Nevertheless, A recent study \cite{barata2021improving} shows that the use of a retrieval network, in a single learner, yields an improvement of 9.2\% in the decision accuracy of dermatologists. Our method indeed suggests that multiple learners capture a data embedding that yields a higher accuracy in clustering and retrieval tasks over single-learner methods, while additionally offering visual saliency from our attention mechanism.

The attention maps produced by our proposed method can serve as proxy pixel-level labels to train a segmentation network. The segmentation results outperform a state-of-the-art method, Attention Residual Learning (ARL) \cite{zhang2019attention}, as well as the recent Embedded Discriminative Attention Mechanism (EDAM) \cite{wu2021embedded} by a margin of 15\% and 17\% in Dice scores, respectively, on the skin lesion dataset. The qualitative results demonstrate that the produced attention maps and their segmentation masks focus on the target lesion, demonstrating the effectiveness and robustness of our method. These attention maps produced in our subspace learning approach could therefore be potentially beneficial to a broader range of weakly supervised tasks, where the feature space remains challenging to represent using a single metric model within a specific task.



\section*{Acknowledgments}
This research work was partly funded by the Canada Research Chair on Shape Analysis in Medical Imaging, the Natural Sciences and Engineering Research Council of Canada (NSERC), and the Fonds de Recherche du Quebec (FQRNT). We would like to acknowledge Compute Canada for providing computing resources used for this work.

\bibliographystyle{IEEEtranE}

\bibliography{biblio}


\end{document}